\documentclass[runningheads]{llncs}
\usepackage{graphicx}
\usepackage{amsmath,amssymb} % define this before the line numbering.
\usepackage{color}
\usepackage{url}

\usepackage{slashbox}
\usepackage{array}
\newcolumntype{L}[1]{>{\raggedright\let\newline\\\arraybackslash\hspace{0pt}}m{#1}}
\newcolumntype{C}[1]{>{\centering\let\newline\\\arraybackslash\hspace{0pt}}m{#1}}
\newcolumntype{R}[1]{>{\raggedleft\let\newline\\\arraybackslash\hspace{0pt}}m{#1}}

\begin{document}
\newcommand{\point}{
	\raise0.7ex\hbox{.}
}

\pagestyle{headings}
\mainmatter

%===========================================================

\title{Deep Architectures for Face Attributes}
\titlerunning{Deep Architectures for Face Attributes}
\authorrunning{Tobi Baumgartner, Jack Culpepper}

\author{Tobi Baumgartner, Jack Culpepper}
\institute{Computer Vision and Machine Learning Group, Flickr, Yahoo, \newline
\texttt{\{tobi, jackcul\}@yahoo-inc.com}}

\maketitle

\begin{abstract}
We train a deep convolutional neural network to perform identity classification using a new dataset of public figures annotated with age, gender, ethnicity and emotion labels,  and then fine-tune it for attribute classification.
An optimal sharing pattern of computational resources within this network is determined by experiment, requiring only 1 G flops to produce all predictions.
Rather than fine-tune by re-learning weights in one additional layer after the penultimate layer of the identity network, we try several different depths for each attribute. We find that prediction of age and emotion is improved by fine-tuning from earlier layers onward, presumably because deeper layers are progressively invariant to non-identity related changes in the input.
\end{abstract}

\section{Introduction}
\label{sec:intro}

We would like to efficiently compute a representation of faces that makes high level attributes, such as identity, age, gender, ethnicity, and emotion explicit.
The deployed system must run within a computational budget, and we would like to maximize prediction accuracy.
Like others, we believe that learning a deep representation that fuses multiple sources of label information is a promising avenue towards achieving this goal~\cite{paper:facial_deep_multi,paper:hyperface,paper:deep_aux}.
In this work, we approach the fusion idea stage-wise, and investigate the problem of how to properly fuse label information from tasks with conflicting invariances.

To illustrate the problem we address, consider the following straight-forward way of jointly solving identity, age, gender, ethnicity and emotion classification.
First, train a deep network to solve a large-scale identity classification task.
The representation learned by this network can then be adapted to another, related task through a process called fine-tuning~\cite{paper:finetune,paper:off_the_shelf,paper:decaf,paper:midlevel}.
To do so, extract features from the penultimate layer of the identity network, and use them to train classifiers for each attribute.
This approach is still commonly used~\cite{paper:attr_wild}, but has a problem: identity is invariant to age and emotion.
Thus, presumably the representation at the penultimate layer of the network trained to solve the identity task will reflect this and impair discrimination of ages and emotions.
To address this, we varied the layer depth from which our feature was extracted and fine-tuned all layers of the network from that point forward, using cross-validation to pick the best depth.
Our experiments show that fine-tuning from the penultimate layer leads to worse results than fine-tuning from an earlier layer onwards.

\begin{figure*}
\centering
\includegraphics[width=240pt]{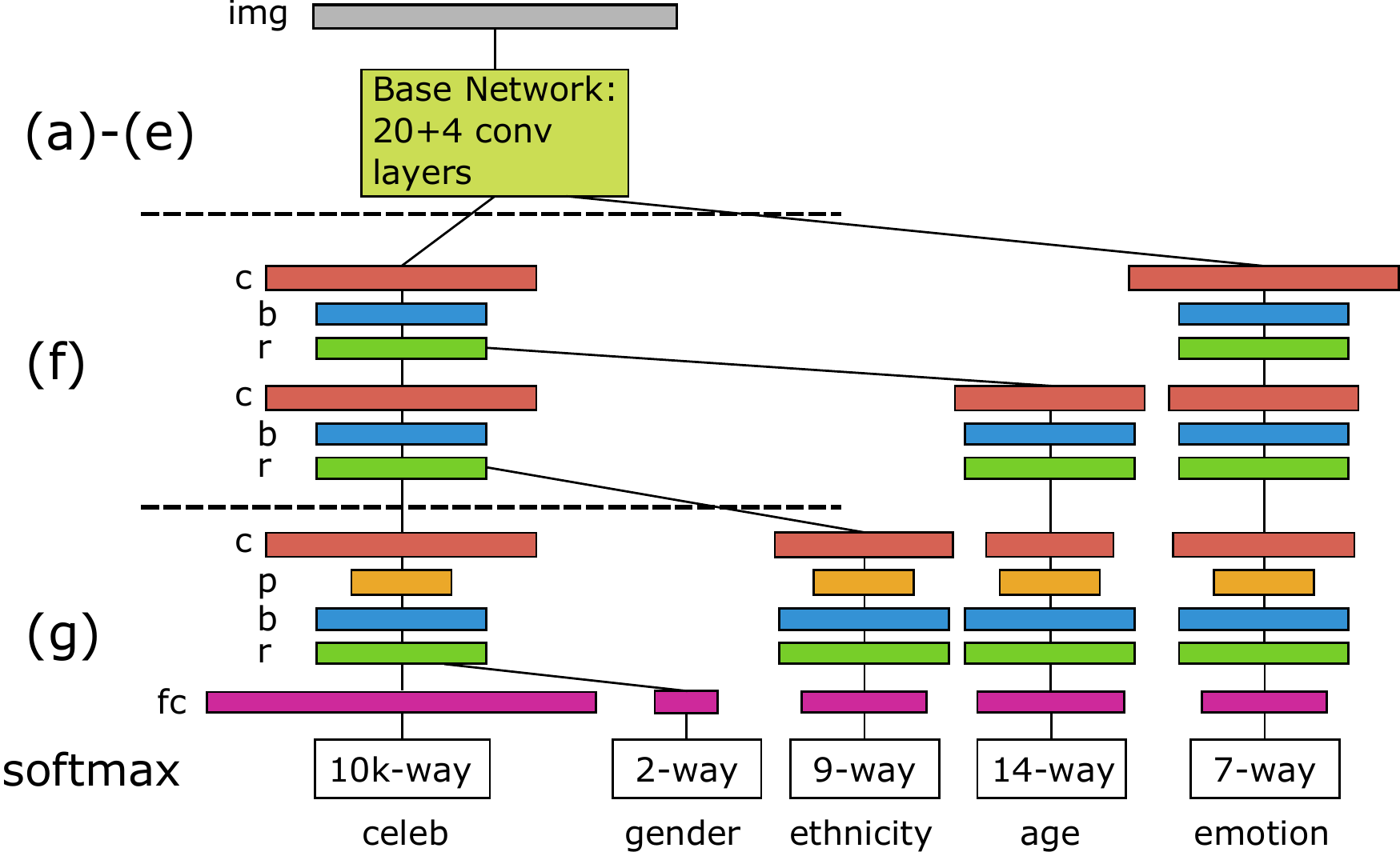}
\caption{Proposed network architecture with fine-tuned layers. The different face attribute sub-task relearn different numbers of top layers. During the forward pass, i.e., in production, we can efficiently compute all outputs simultaneously (cf. Fig.~\ref{fig:jacknet} for full network and legend).}
\label{fig:fine_tune}
\end{figure*}

Our approach has several advantages over the more typical approach described above.
A significant amount of the identity network is shared across attribute predictors, yet the invariance conflicts between tasks are resolved.
Our approach is pragmatic when it comes to dealing with a high variance in the number of labels from one task to another -- due to cross validation, tasks with more labels will benefit from them by sharing fewer weights with other tasks.

Fig.~\ref{fig:fine_tune} shows the architecture that we derive from experimentation. It shows the branching pattern that gives the best prediction accuracy for each task, and the order of depths gives a measure of the relationship of the task to identity classification. Admittedly, this relationship is slightly confounded by the variable number of training examples across tasks.

\section{Related Work}
\label{sec:relatedwork}

\paragraph{\textbf{Fusion of multiple information sources.}}
Liu \textit{et al.} jointly fine-tune a location specialized network and a network for attribute detection and show that fusing information from attribute labels yields an improvement in the localization task~\cite{paper:attr_wild}. Their attribute network (ANet) is pretrained using a massive identity database, and they use the representation from the penultimate layer to learn attribute predictors such as ``is wearing sunglasses". If their identity database contained any individual pictured both with and without sunglasses, then their pretraining would have encouraged the penultimate layer to be invariant to this attribute. They did not utilize the technique we describe herein, which suggests that it is non-obvious and deserving of some attention.

Zhang \textit{et al.} show that face attribute labels can improve landmark detection~\cite{paper:facial_deep_multi}. They simultaneously train with multiple attributes (\textit{wearing glasses, smiling, gender, pose}) and landmark labels and introduce an early stopping criteria for each of the tasks. They use the same shared base feature computation for all tasks, but also branch only at the last fully connected layer.

Surprisingly little attention has been given to the topic of network architectures with weight sharing patterns that provide benefits to multiple tasks with conflicting invariances such as ours. The idea of branching at different layers of a deep network for different tasks is utilized in~\cite{paper:abhinav}, but their branching pattern is not evaluated quantitatively.

Fine-tuning has variously been shown to be a very effective and powerful tool to learn a specialized task on low amounts of data. Razavian \textit{et al.}~\cite{paper:off_the_shelf} and Donahue \textit{et al.}~\cite{paper:decaf} amongst others~\cite{paper:deepvis,paper:midlevel} show that standard CNNs trained on ImageNet~\cite{paper:imagenet} can be adapted to perform attribute classification. Zhang \textit{et al.} infer various human attributes by training a number of part-based models with pose-normalized CNNs~\cite{paper:panda}.

\paragraph{\textbf{Face attribute datasets.}}
Research in \textit{emotion recognition} is supported by 3 standard benchmarks. EmotiW~\cite{paper:emotiw} is a labeled set of movie clips, in which each clip portrays one of 7 emotions; hence, even though the dataset is fairly large with $400k$ images, it is not very diverse. In the MultiPie~\cite{paper:multipie} dataset, a set of only 4 facial expression, as well as facial landmarks, are captured in very high detail with 15 cameras.
The Toronto Face Database (TFD) by Susskind \textit{et al.}~\cite{paper:torontoface} contains 7 emotions on a few thousand images.

The task of \textit{age recognition} is addressed in the adience dataset by Levi \textit{et al.}~\cite{paper:adience} and the cross-age celebrity dataset (CARC) by Chen \textit{et al.}~\cite{paper:carc}. The former uses the yfcc dataset\footnote{\url{http://webscope.sandbox.yahoo.com/catalog.php?datatype=i&did=67}} and has $19k$ images. The latter implies celebrities' ages from photo timestamps and consists of $160k$ images.

To the best of our knowledge, there are no widely available diverse datasets for the task of \textit{ethnicity recognition}. 

The above datasets contain images from various ages/genders/ethnicities, but they all have limitations which make them unsuitable for our purposes: LFW and PubFig83 are too small; EmotiW, MultiPie, and Gallagher do not have enough diversity; EmotiW, TFD, and CARC are only available for academic use.

\section{Model}
\label{sec:model}

\begin{figure}
\centering
\includegraphics[width=180pt]{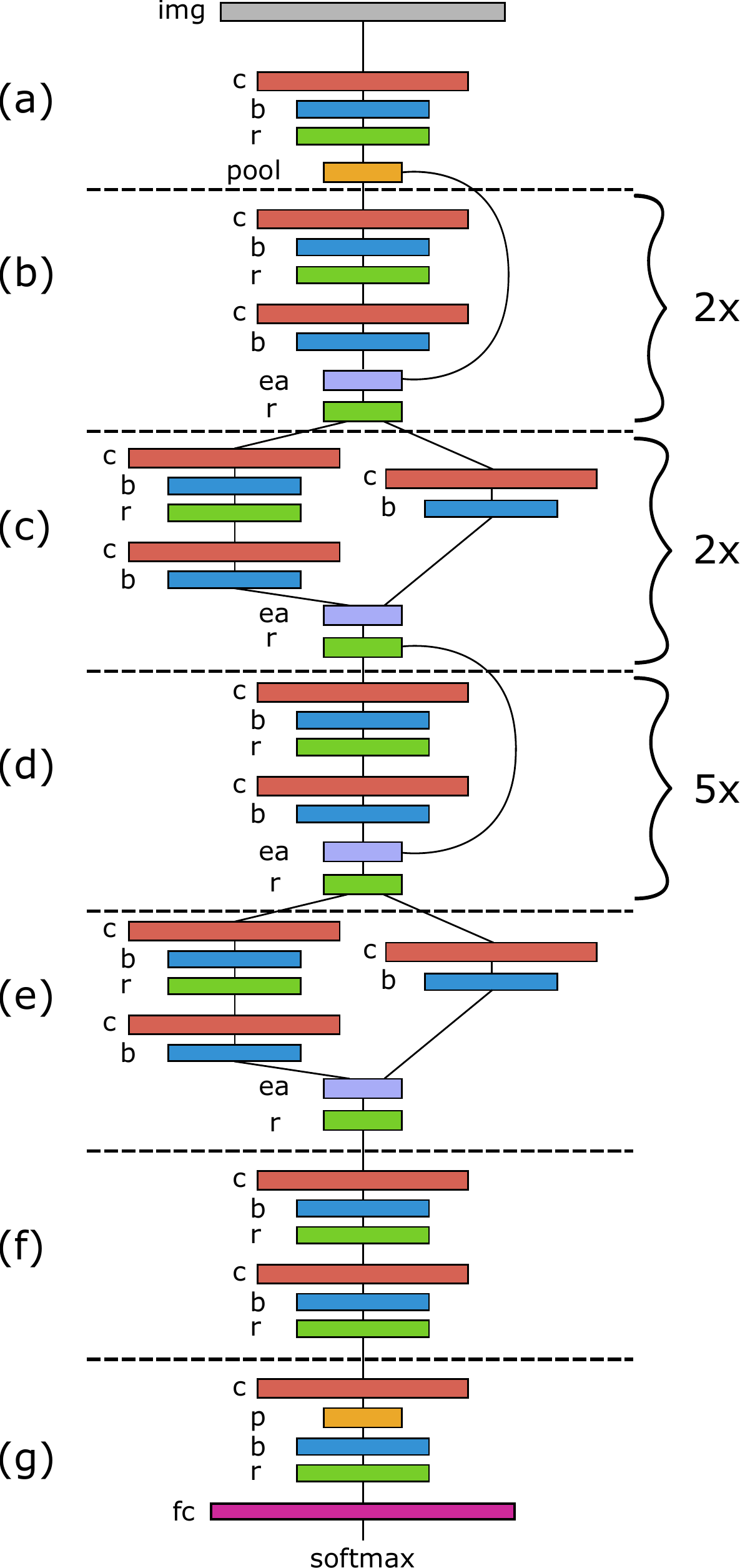}
\caption{The complete face-network. c=convolution, b=batchnorm, r=ReLU, ea=elementwise add, p=pool, fc=fully connected. Parts (b), (c) and (d) are repeated as indicated. Excluding the shortcut layers, there are 24 convolution layers.}
\label{fig:jacknet}
\end{figure}

In the following we describe our base network architecture for identity recognition, as displayed in Fig.~\ref{fig:jacknet}, as well as the fine-tuned top layers, shown in Fig.~\ref{fig:fine_tune}.

\subsection{Base Network}
We first train a deep convolutional neural network (CNN) to solve the task of celebrity recognition, then fine-tune the upper layers of this network to solve the gender, ethnicity, age, and emotion tasks.
The design for our CNN is inspired by the recent work on \textit{residual networks}~\cite{paper:resnet}. Like the networks in that paper, ours is built from a few simple functions: convolution, ReLU, pooling, and batch normalization. We compose them together to build a complex, high capacity function with about 10M free parameters, using skip-connections to mitigate the degradation problem. 
We use the softmax function to convert the outputs of the topmost layer into a probability distribution and minimize the cross entropy between this and the label distribution. The total loss is the cross entropy averaged across all samples.

Fig.~\ref{fig:jacknet} shows the structure of our base network. The input is a 224x224 pixel RGB image resulting from the output of our detection and alignment algorithms. In stage (a) we convolve 32 7x7 filters with stride 2 then apply 2x2 max pooling with stride 2; this quickly reduces the spatial resolution of the input to 56x56. From this point forward, we repeat the basic module shown in (b): a dimensionality reducing 1x1 convolution, followed by batch normalization and ReLU, followed by a dimensionality expanding 3x3 convolution, followed by element-wise addition with a shortcut connection. Stages (c) and (e) reduce the spatial resolution of the input by performing their 3x3 convolutions with stride 2. These reductions in the spatial dimensions are accompanied by a corresponding factor of 2 increase in the number of filters in the 3x3 convolutional layer, and the shortcut connection uses 1x1 convolutions to equalize dimensions, also at stride 2. The bulk of the flops and parameters are in stage (d), which is a succession of 1x1 convolutions to reduce the dimensionality to 128, followed by 3x3 convolutions that increase it to 256. Stage (d) operates at 14x14 spatial resolution. Stage (f) increases the dimensionality to 512, and stage (g) uses average pooling to collapse the remaining 7x7 spatial dimensions, and imposes a 320 dimensional bottleneck immediately preceding the final fully connected layer. In total, the network requires 0.9 Gflops to process an input image, and has 9M free parameters.

\begin{figure}[t!]
\centering
\renewcommand{\arraystretch}{1.15}
\begin{tabular}[center]{l|C{1.75cm}|C{1.75cm}|C{1.75cm}|C{1.75cm}}
\backslashbox{layer}{attribute} & \textbf{emotion} & \textbf{age} & \textbf{ethnicity} & \textbf{gender} \\
\hline
\texttt{conv17} & 0.44 & 0.22 & 0.60 & 0.98 \\
\texttt{conv19} & \textbf{0.68} & 0.28 & 0.68 & 0.94 \\
\texttt{conv21} & 0.30 & 0.24 & 0.50 & 0.98 \\
\texttt{conv22} & 0.34 & \textbf{0.40} & 0.55 & 0.95 \\
\texttt{conv-bn320} & 0.42 & 0.06 & 0.61 & 0.97 \\
\texttt{fc} & 0.44 & 0.24 & \textbf{0.72} & \textbf{0.99}
\end{tabular}
\caption{Grid of attribute fine-tuning experiments. Accuracy of face attribute tasks on validation set after convergence.}
\label{tab:cross_exps}
\end{figure}

\subsection{Fine Tuning}
\label{sec:fine_tuning}

When trained discriminatively using large datasets, deep networks learn representations of the data that can be re-purposed for related tasks through a process called fine tuning~\cite{paper:finetune,paper:off_the_shelf,paper:decaf,paper:midlevel}. After the basic network in Fig.~\ref{fig:jacknet} is fully trained, we fine-tune by replacing the top layers of the network, then use our datasets for the tasks of gender, age, ethnicity and emotion recognition to optimize only the weights in these new layers. To be clear, in these new layers, only the number of units in the final layer changes -- it is set according to the number of target classes in the task. The other layers retain the architecture of the base network, but the weights in the new layers are initialized randomly and retrained. We depict this in Fig.~\ref{fig:fine_tune}. Stages (a)-(e) are summarized in the `base network' box. We replace the $10k$-celebrity softmax layer (`fc'-box in Fig.~\ref{fig:fine_tune}(g)) with a task-specific softmax, and set different top-layers according to the depiction in Fig.~\ref{fig:fine_tune}. Once trained, the parameters of the base network are fixed and not altered during fine-tuning. 

The task-specific architectures shown in Fig.~\ref{fig:fine_tune} were determined experimentally, using a validation set. In Tab.~\ref{tab:cross_exps} we show the results after fine-tuning for our tasks for a variety of branch depths. The cross-validation results give an ordered relationship between tasks, based on the branching depth: \textit{emotion} branches earliest, then \textit{age}, \textit{ethnicity}, and, finally, \textit{gender} performs best when we branch off the penultimate layer, as is typical.

Fine-tuning the network from earlier layers than the ones experimentally determined yields inferior results, likely due to overfitting caused by the limited number of labeled training samples for the attribute tasks. There is a trade-off between overfitting (backing off too far) and being too invariant to learn the task (not retraining enough layers of the network).

The depth of the best performing connectivity pattern has a noteworthy relationship to the invariances learned by each layer in the base network: due to the way it was trained, higher layers should be progressively more invariant to changes in the input that are unrelated to changes in the identity of the person pictured. That is, it makes perfect sense that the information required to discriminate emotional states should be unavailable at the top of the network -- if it were, it would only degrade the capability of the network to identify a celebrity. Put another way, person identity is invariant to changes in emotion or age, so the features that are predictive of these concepts are found lower in the network, before the network has learned to distinguish identities. 
Gender and ethnicity are most closely related to the identity of a person, i.e. a person has a fixed ethnicity and an (almost) fixed gender, so it is natural that gender/ethnicity prediction relies on features that are more closely associated with identity.

Sharing the representation among related tasks significantly reduces the computational complexity. This allows us for fast evaluation of all tasks at test time in our production system, since the forward-pass through the base network only has to be computed once.
Tab.~\ref{tab:data} summarizes the data and number of fine-tuned parameters for the specific tasks. In the following we describe the data collection, training process, and the evaluation results for each of the 5 tasks in more detail.

\section{Experiments}
In this section we show experimental results for our deep network. The applications and experiments in Sec.~\ref{sec:ident} uses the entire network and resulting feature vectors. In Secs.~\ref{sec:gender}, \ref{sec:age}, \ref{sec:emotion}, and \ref{sec:ethnicity}, we fine-tune the network for face attribute specific tasks. In Sec.~\ref{sec:cross_val} we propose an interpretation of our findings and relate them to the capacity of the deep net to learn levels of invariance.

\subsection{Identifying Public Figures}
\label{sec:ident}
In order to learn a mapping from pixels to a representation that is invariant to changes in the pixels that are unrelated to identity, we use a large dataset with many examples of many identities. As there was no suitable, freely available dataset, we collected our own proprietary dataset by analyzing the Yahoo image search engine query logs.

We started with a list of 40k public figures from wikipedia~\footnote{\url{http://wikipedia.org}} and identified those for which we had at least 50 clicked images in the search logs (i.e. at least 50 positive training examples). For unbiased evaluation of our algorithm against the LFW benchmark, we also excluded identities that appear in LFW, resulting in about 10K identities and a total of 2.5M images.

We partitioned the dataset into training and validation, keeping 5 samples from each identity for validation. We initialize all weights in the network by drawing from a Gaussian distribution with mean $0$ and variance $0.01$, and optimize the cross entropy loss using stochastic gradient with momentum 0.9 and a batch size of 400. Initially, our learning rate is $0.1$ and we reduce it by a factor of 4 every 10,000 minibatches. After training, our validation recognition accuracy on the 10K-way classification task is 84\%.

\subsection{Face Verification}
\label{sec:verification}

In the LFW task, pairs of faces must be `verified' as belonging either to the same person, or to two different people~\cite{paper:lfw}. The dataset contains 1,680 people and 6,000 pairs of faces arranged in 10 splits with 300 `same' and 300 `not same' pairs each. We evaluated the `unrestricted' protocol using the following leave one out evaluation paradigm. For each of the 6,000 pairs of faces, we run our detection and alignment, then extract the 320 dimensional features from the fully connected layer in~\ref{fig:jacknet}(g). We calculate the cosine similarity between each pair of 320d vectors. We select the threshold that provides the best verification on 9 splits, then evaluate it on the remaining one. We reach an accuracy of 95.98\% on LFW.

\begin{figure}[t]
\centering
\begin{tabular}[center]{l|l|c|c|c}
\textbf{task} & \textbf{size} & \textbf{accuracy} & \textbf{n-way} & \textbf{\#params}\\
\hline
celebrity & 2.5M & $84\%$ & $10k$ & $10M$ \\
% 131,072 + 393,216 + 491,520 + 320*7 + 7 
emotion & 280k & $68\%$ & 7 & $1,018,055$\\ 
% 393,216 + 491,520 + 320*14 + 14
age & 290k & $40\%$ & 14 & $889,230$\\
% 491,520 + 320*9 + 9 
ethnicity & 197k & $72\%$ & 9 & $642$\\
% 320*2+2 
gender & 6M & $99\%$ & 2 & $642$ \\
\end{tabular}
\caption{This table summarizes the size of our newly collected datasets and the accuracy we achieve on this data. Each of the tasks has a different complexity, that can be described by the n-way classification. The trained layers for each task are illustrated in Fig.~\ref{fig:fine_tune}. The last column shows the number of free parameters that are learned during fine-tuning.}
\label{tab:data}
\end{figure}

\subsection{Attribute Tuning}
\label{sec:cross_val}
After the base network is fully trained and reaches an accuracy of 84\% on the 10k-way classification task, we fine-tune it starting at different layers for the 4 attribute tasks. For this, we initialize the same network with the pretrained weights and freeze the weights before the respective starting layer; during back-prop these weights will not be adjusted. We also replace the last fully connected layer with a slimmer fully connected layer that has the number of targets of the specific task. In Tab.~\ref{tab:data} we show the number of labels for each of the attributes. All of the experiments have a 1-hot classification label and we summarize the accuracy on the test set in Tab.~\ref{tab:cross_exps}.

Gender and ethnicity prediction are closely related to identity prediction, and are well-suited to be learned by fine-tuning from the penultimate layer of an identity discriminating network. Age and emotion prediction benefit from branching earlier and retraining more layers. Furthermore, our results in Tab.~\ref{tab:cross_exps} imply that age is more closely related to identity than emotion, which is likely explained by the prevalence in our identity database of photos from a certain time period in a public figure's life.

\subsection{Predicting Gender}
\label{sec:gender}
For \textit{gender classification} we can bootstrap the aforementioned celebrity dataset; The gender of each celebrity can be extracted from the Yahoo Knowledge graph. In order to avoid overfitting, we separate the training and testing sets by celebrities, i.e., the same person will not appear in both the training and test sets. For gender prediction, we achieve accuracy of 99\% on a dataset of 6M faces (note that the dataset is larger than for the 10k-way classification task as we do not need to perform the same filtering to have 50 training examples per celebrity).

\subsection{Predicting Age} 
\label{sec:age}
Similarly to the gender dataset, we bootstrap the data for \textit{age recognition} from our celebrity dataset, again by using the Yahoo Knowledge graph, in this case to determine the birthdays of every celebrity. We then determine the capture date of every downloaded celebrity image from the exif-data, if available. Although this process is subject to some noise in the label collection, a similar approach was used in the Cross-Age Reference Coding benchmark~\cite{paper:carc}.

Since not all the celebrity images have valid exif-dates, we end up with a dataset for age recognition of about $290k$ images. We split this data into 14 age bins, similar to the adience dataset~\cite{paper:adience}. With this, instead of perfectly predicting a persons age, the task now becomes to predict an age range. The adience dataset contains $19k$ images and Eidinger \textit{et al.}~\cite{paper:adienceres} achieve an accuracy of $45.1\%$. In their experiment, they only consider 8 age-bins. The binning itself is the same as in our dataset, but they leave out intermediate bins, like 21-24. In our experiments we achieve and accuracy of $40\%$.

\subsection{Predicting Emotion}
\label{sec:emotion}

For \textit{emotion recognition}, we collect a new set of public images from Flickr~\footnote{\url{http://flickr.com}} that have a high probability of containing emotions. We first filter the $15B$ images in the Flickr corpus by the autotag "portrait". This tag has been computed using the face detector described in Sec.~\ref{sec:model}; which adds this tag if exactly 1 face is detected in an image and the size of the face is larger than $0.3$ times the image height and width. We find all publicly available images that have a emotion-related user-generated tag or title associated with it. Our system distinguish between the following 7 emotions: happy, sad, angry, surprised, disgusted, scared and neutral. For each of these emotions, we compile a list of synonyms and facial expressions and arrive at a total set of 200+7 terms for collecting user-labeled data. This emotion dataset consists of $280k$ images. Our best results yield a recognition accuracy of $68\%$. We also compare the performance of our network to the EmotiW~\cite{paper:emotiw} challenge, for which Kahou \textit{et al.}~\cite{paper:emonet} reported accuracy of $41.03\%$ in the 2013 challenge on the task of emotion detection from images. With our method we achieve an accuracy of $30.23\%$ on their data. As already mentioned in Sec.~\ref{sec:relatedwork}, the EmotiW data does not reflect a real-life task as presented in our scenario though.

\subsection{Predicting Ethnicity}
\label{sec:ethnicity}
The task of \textit{ethnicity detection} is not strictly well defined. There are different levels of granularity imaginable, e.g. Hispanic vs. Mexican, White vs. Central European. In this example, one term is based on region and the other on country borders.
Ethnic groups identify by varying commonalities among their peers instead of just the color of their skin. These unifying characteristics are as diverse as: region (South Islanders), language (Gaels), nationality (Iraqis), ancestry (Afro-Brazilian) or religion (Sikhs)\footnote{\url{https://en.wikipedia.org/wiki/Ethnic_group}}.

We adapt our celebrity dataset for this task by extracting celebrity ethnicity labels from \texttt{EthniCelebs}~\footnote{\url{http://www.ethnicelebs.com}}, which has a  well curated set of ethnicity labels for over $15k$ people. We use this data to enhance our celebrity dataset with their family origins. The resulting labels are very specific, e.g., \textit{austro-hungarian jewish} or \textit{ulster-scots}. We next map these locations to 9 ethnicities in accordance with industry competitor Getty~\footnote{\url{http://www.gettyimages.com/}}. The resulting ethnicities we learn to distinguish between are: black, hispanic, east-asian, middle-eastern, pacific islander, south-asian, south-east-asian and white. Our dataset has a size of $197k$ images and we achieve an accuracy of $72\%$.

Obviously, the ethnicity of most people cannot be described in a single word. We have different backgrounds with parents from various places. To reflect this, we also trained a multi-target classifier for ethnicity detection; This means that the result of this model will be a 9-dimensional vector with probabilities that the tested person belongs to each of these ethnic groups. The evaluation of this task is done in terms of \textit{true/false positive rate} (\textit{tpr/fpr}). The \textit{tpr} describes the percentage of actual ethnicity, that we correctly predict; a non-perfect score means that we miss to label a person with an ethnicity they identify with. More important for this sensitive task is the \textit{fpr} though, since it measures the number of times we falsely assign an ethnicity to someone. People might take offense in errors of this kind. In our experiments, we chose an operating point with a \textit{tpr} of $71.97\%$ and \textit{fpr} of only $1.03\%$; we err on the safe side, while still assigning some label to almost all test subjects: only $0.44\%$ do not get a ethnicity prediction at all.

\section{Discussion}

The representations learned by deep networks trained in an identity discrimination task can be fine-tuned to perform well in the tasks of age, gender, ethnicity and emotion recognition. We expose limitations in the typical approach, and show benefits to careful architecture analysis via cross-validation.
Our approach provides a straight-forward way to measure the level of similarity between two tasks, in terms of where the information required to solve them is represented inside the network.
Capturing these ideas in a learning approach remains for future work.

\section*{Acknowledgments}
We would like to thank Neil O'Hare for collaborating with us on the search engine query logs, and the entire Yahoo Vision and Machine Learning Team.

\bibliographystyle{abbrv}
\bibliography{faceattr}

\end{document}